\documentclass[letterpaper, 10 pt, conference]{ieeeconf}  

\IEEEoverridecommandlockouts                              

\usepackage{url}
\usepackage{cite}
\usepackage{amsmath,amssymb,amsfonts}
\usepackage{algorithmic}
\usepackage{graphicx}
\usepackage{textcomp}
\usepackage{xcolor}
\usepackage{multirow}
\usepackage{subcaption}

\usepackage{dblfloatfix}    

\title{\LARGE \bf
NanoFlowNet: Real-time Dense Optical Flow on a Nano Quadcopter
}

\author{Rik J. Bouwmeester, Federico Paredes-Vallés and Guido C. H. E. de Croon
\thanks{All authors are with the Micro Air Vehicle Laboratory, Faculty of Aerospace Engineering, Delft University of Technology, Delft, The Netherlands. Contact: {\tt\small f.paredesvalles@tudelft.nl}}
}

\begin{document}

\captionsetup{font=footnotesize}
\captionsetup[sub]{font=footnotesize}

\maketitle
\thispagestyle{empty}
\pagestyle{empty}

\begin{abstract}

Nano quadcopters are small, agile, and cheap platforms that are well suited for deployment in narrow, cluttered environments. Due to their limited payload, these vehicles are highly constrained in processing power, rendering conventional vision-based methods for safe and autonomous navigation incompatible. Recent machine learning developments promise high-performance perception at low latency, while dedicated edge computing hardware has the potential to augment the processing capabilities of these limited devices. In this work, we present NanoFlowNet, a lightweight convolutional neural network for real-time dense optical flow estimation on edge computing hardware. We draw inspiration from recent advances in semantic segmentation for the design of this network. Additionally, we guide the learning of optical flow using motion boundary ground truth data, which improves performance with no impact on latency. Validation results on the MPI-Sintel dataset show the high performance of the proposed network given its constrained architecture. Additionally, we successfully demonstrate the capabilities of NanoFlowNet by deploying it on the ultra-low power GAP8 microprocessor and by applying it to vision-based obstacle avoidance on board a Bitcraze Crazyflie, a 34 g nano quadcopter.

\end{abstract}

\section{INTRODUCTION}

Safe and reliable navigation of autonomous aerial systems in narrow, cluttered, GPS-denied, and unknown environments is one of the main open challenges in the field of robotics. Because of their small size and agility, micro air vehicles (MAVs) are optimal for this task \cite{Floreano2015, Bodin2018}. Nano quadcopters are a variety of MAVs that are characterized by minimal weight (approx. 30 g) and size (approx. 10 cm rotor-to-rotor) and hence are well suited for deployment under the aforementioned conditions. With the right algorithm design, these nano quadcopters have been demonstrated to be able to perform complex tasks such as exploration \cite{McGuire2019a} or gas source seeking \cite{Duisterhof2021}. However, conventional approaches to these problems rely on computationally expensive ``map-based'' methods that require an array of sensors (e.g., stereo camera, LiDAR) and processors that, in the majority of cases, exceed the payload capacity of these vehicles.

The main approach to autonomous flight of MAVs is based on monocular vision, since a single camera can be light-weight and energy-efficient, while providing rich information on the environment. One of the most important monocular visual cues for navigation is optical flow. Until now, it has been extensively exploited on aerial vehicles with relatively high payload capacity for tasks such as obstacle avoidance \cite{Gao2017, Sanket2018}, and several bio-inspired methods for autonomous navigation \cite{Conroy2009, Zingg2010, DeCroon2016, Serres2017, DeCroon2021}.

\begin{figure}
    \centering
\begin{subfigure}[t]{0.45\textwidth}
    \includegraphics[width=\textwidth]{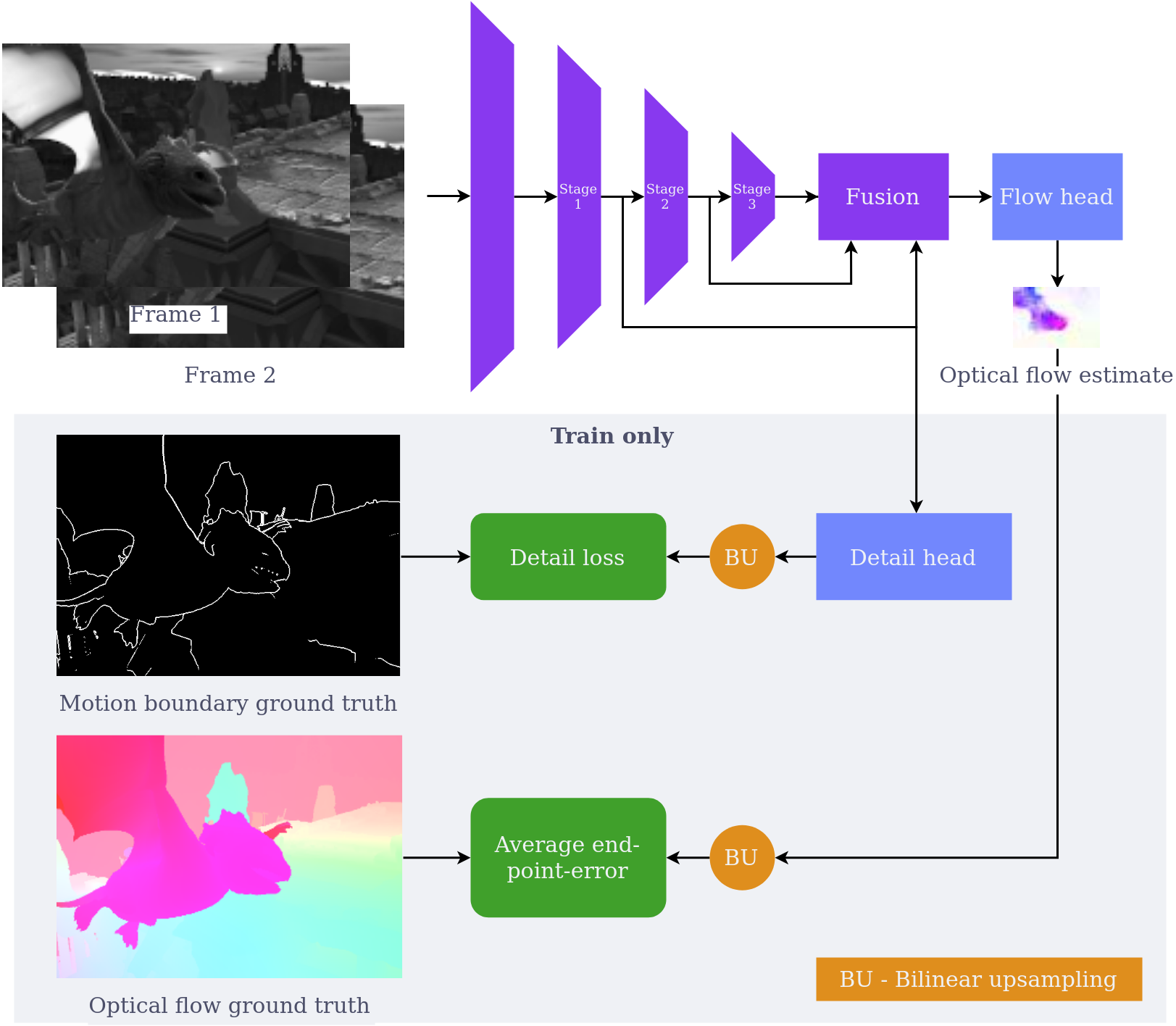}
\end{subfigure}\vspace{10pt}\hfill\vfill
\begin{subfigure}[t]{0.45\textwidth}
    \includegraphics[width=\textwidth]{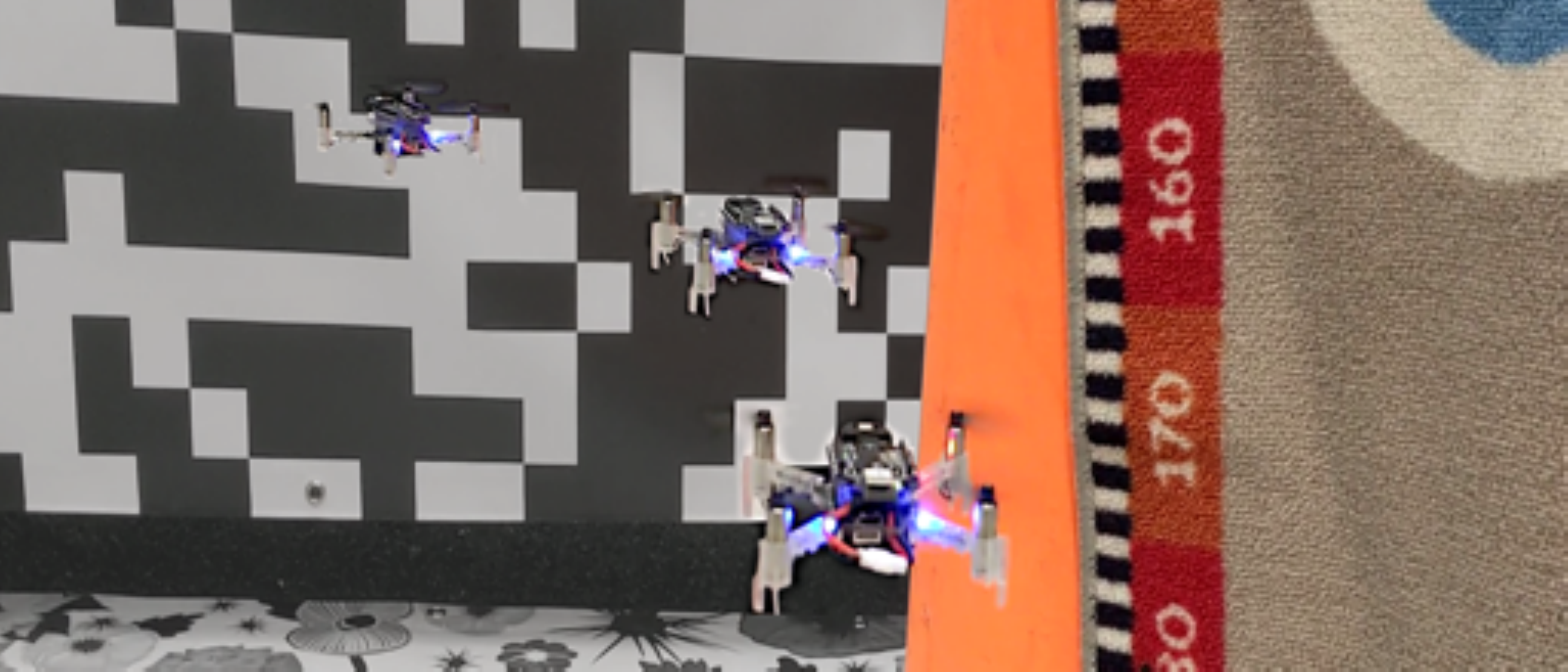}
\end{subfigure}
\caption{\textit{Top}: NanoFlowNet consists of (i) an encoder that extracts features from the 
input images, (ii) a fusion module that combines features from different levels, and (iii) a motion-boundary-guided detail head, which is only enabled during training, to guide the learning with zero cost to inference latency. \textit{Bottom}: We demonstrate NanoFlowNet in an obstacle avoidance application on board a nano quadcopter (time-lapse image).\vspace{-0pt}}
\label{fig:avoidanceexample}
\end{figure}

Traditionally, the task of monocular optical flow estimation has been performed by hand-crafted methods \cite{Lucas1981, Horn1981b}. However, the field recently shifted toward deep learning approaches \cite{Dosovitskiy2015, Ranjan2017, Ilg2017, Zhao2017, Sun2018, Hui2018a, Yin2019, Yang2019, Hui2020a, Hui2020, Zhao2020, Teed2020}, which deliver not only a better performance than the conventional methods but also a faster runtime. Although the focus has largely been on improving performance, efforts have been made to find models of reduced size and faster inference \cite{Ranjan2017, Hui2018a, Hui2020a, Hui2020, Sun2018, Hur2019, Ilg2017}. However, these methods remain computationally expensive, with runtime ranging from several to tens of frames per second (FPS) on desktop GPUs and requiring millions of parameters (and hence large amounts of memory), rendering these models incompatible with edge hardware.

In this work, instead of improving the accuracy of state-of-the-art approaches, we focus on their inference speed and, more particularly, on the deployment of a dense optical flow network on edge devices. To this end, we present \textit{NanoFlowNet}, a lightweight convolutional neural network (CNN) architecture for optical flow estimation that, inspired by the semantic segmentation network STDC-Seg \cite{Fan2021}, achieves real-time inference on the ultra-low power GAP8 multi-core microprocessor on the Bitcraze AI-deck. An overview of the proposed network architecture and its training pipeline can be found in Fig. \ref{fig:avoidanceexample}.

The key contributions of this paper are listed as follows. First, we introduce NanoFlowNet, a novel lightweight neural network architecture that performs, for the first time, real-time dense optical flow estimation on edge hardware. We validate this network, which runs at 5.5-9.3 FPS on the tiny GAP8 microprocessor, through extensive quantitative and qualitative evaluations on multiple datasets. Second, we show, for the first time, that using motion boundary ground truth to guide the learning of optical flow improves performance while having zero impact on inference latency. Last, we demonstrate the proposed NanoFlowNet in a real-world obstacle avoidance application on board a Bitcraze Crazyflie nano quadcopter.

The remainder of this paper is organized as follows. Section II provides an overview of the state-of-the-art on autonomous navigation of nano quadcopters and on real-time inference with CNNs. In Section III, we present the details of the proposed architecture and its training pipeline, while Section IV covers the setup of the experiments as well as the obtained results. Finally, concluding remarks are given in Section V alongside recommendations for future work.

\section{Related Work}
\subsection{Autonomous navigation of nano quadcopters}

The limited computational capacity of nano quadcopters (and MAVs in general) puts a constraint on the types of methods that can be used for autonomous navigation. Methods demonstrated on board nano quadcopters can be broadly grouped in model-based reinforcement learning for hovering \cite{Lambert2019a}, obstacle avoidance based on dedicated laser ranging sensors \cite{McGuire2019a, Duisterhof2020, Duisterhof2021}, and self-motion estimation using optical flow from dedicated optical flow sensors \cite{Briod2013} or estimated with external, multi-camera setups \cite{Moore2014, McGuire2017}. Other methods circumvent the computational constraints of these vehicles by running methods off-board \cite{Dunkley2014, Candan2018a, Anwar2019}. 

Regarding edge computing hardware, recent works have focused on augmenting the computational power of nano quadcopters without exceeding their payload limitations. Methods based on application-specific integrated circuits (ASICs) \cite{Suleiman2019, Li2019a, Hosseini2021, Manjunath2021} can efficiently provide information for specific tasks such as SLAM and visual-inertial odometry but have not yet been presented on a flying drone. More recently, parallel ultra-low power processors introduce energy-efficient multi-core processing to parallelize visual workloads on edge devices \cite{Palossi2019a}. In this work, we exploit the commercially available off-the-shelf AI-deck from Bitcraze, equipped with the GreenWaves GAP8 system-on-chip (SoC) and an ultra-low power grayscale camera. This nine-core SoC has been used for several end-to-end methods that integrate perception and navigation by directly regressing inputs through a CNN into control commands \cite{Palossi2019a, Palossi2019, Palossi2021}. Instead, in our approach, we calculate optical flow as an intermediate step. This gives us direct control over vehicle behavior and can support multiple optical-flow-based tasks to be performed simultaneously or interchangeably. Our work, motivated by these benefits, is the first to present a fully convolutional neural network for a dense (i.e., per-pixel) prediction task on board the AI-deck.

\subsection{Real-time dense inference with CNNs}


\begin{figure*}[b]
    \centering
    \includegraphics[width=0.85\textwidth]{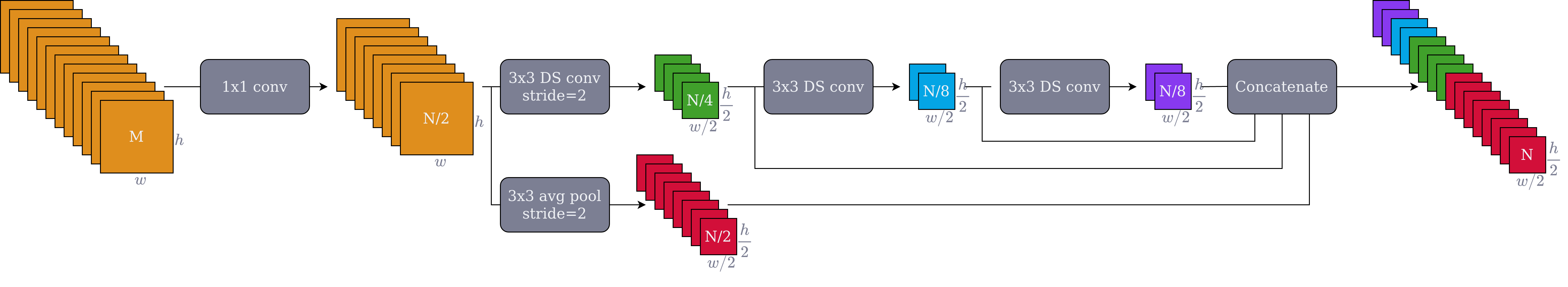}
    \caption{Original strided STDC module from \cite{Fan2021}, with the exception that we use depthwise separable (DS) convolutions in place of all non-pointwise convolutions. We use ReLU activations after all layers in the block. M denotes the number of input features, while N is the number of output features.}
    \label{fig:stdc_original}
\end{figure*}

\begin{figure*}[b]
    \centering
    \includegraphics[width=0.85\textwidth]{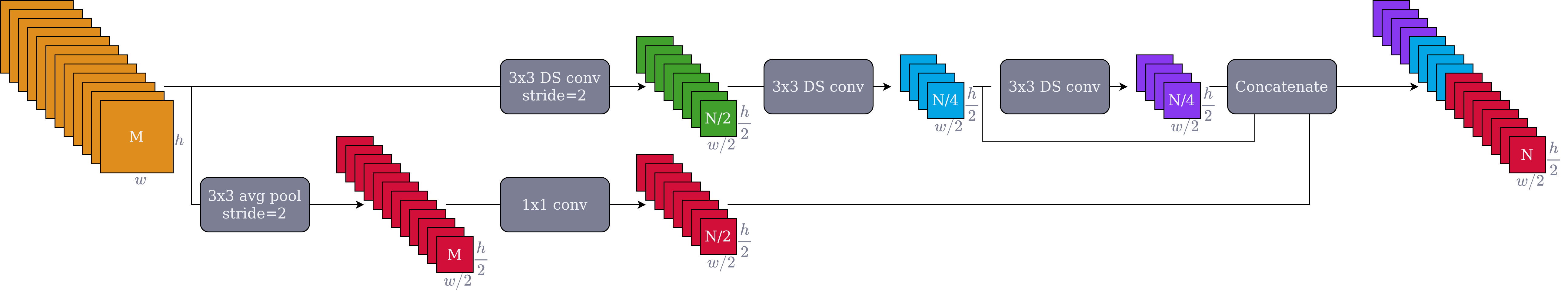}
    \caption{Our modified strided STDC module. We reorganize the operations to minimize the spatial resolution pointwise convolutions have to perform on.}
    \label{fig:stdc_modified}
\end{figure*}

For the design of NanoFlowNet, we draw inspiration from recent semantic segmentation literature in order to significantly speed up optical flow estimations while retaining performance. More specifically, we draw inspiration from the BiSeNet \cite{Yu2018} and STDC-Seg \cite{Fan2021} architectures. First, BiSeNet identified a sacrifice of low-level spatial information in previous real-time methods and improved performance by proposing a multi-path architecture in which low-level spatial information is encoded in a separate path. A feature fusion module was proposed to fuse information from the high- and low-level paths, while an attention refinement module refined features through channel attention. Then, the STDC-Seg architecture introduced the STDC module, which increases the receptive field size per layer at a low computational cost. Furthermore, it identified that BiSeNet's spatial path pronounces edges, and replaced the convolutions from the path with a train-time-only ``detail head'' and ``detail loss'' to mimic the information passed from the removed convolutions, thus shrinking the model and decreasing latency. The ``detail guidance ground truth'' was generated by convolving the ground truth segmentation map with a Laplacian kernel.

A few elements of STDC-Seg and BiSeNet have been separately investigated in the context of optical flow. AD-Net \cite{Zhai2019} showed that channel attention can be beneficial for optical flow estimation, while EDOF \cite{Zuo2021} fused features from an edge-detector network and an optical flow encoder network for detail-guided optical flow estimation. Similar to STDC-Seg, we use edges to guide the learning.

\section{Method}

For the design of NanoFlowNet, we adopt the STDC-Seg network \cite{Fan2021} and modify it to our needs. We replace all regular convolutions with depthwise separable convolutions, and we globally reduce the number of filters by a factor of four to further reduce latency and the number of parameters. We introduce an even smaller model with half of NanoFlowNet's filters (globally) and call it \textit{NanoFlowNet-s}. Further modifications to the architecture are discussed in detail in the following subsections. 

\subsection{Motion boundary detail guidance}
The closest analogy to detail guidance as used in STDC-Seg is to generate edges from the optical flow ground truth. Instead, we replace this ``edge-detect" detail guidance ground truth with motion boundary ground truth from the optical flow datasets. 
We adopt the Focal Loss \cite{Lin2017} to counter the class imbalance problem.

\subsection{Strided STDC module redesign}

We modify the strided STDC modules from STDC-Seg \cite{Fan2021} to further decrease latency. The original and modified strided STDC modules can be found in Figs. \ref{fig:stdc_original} and \ref{fig:stdc_modified}, respectively. First, following the insights of several low-latency literature methods \cite{Chollet2016, Howard2017, Zhang2017, Sandler2018, Bazarevsky2019}, we replace all convolutions in the STDC module with depthwise separable convolutions due to their low computational expense. Second, we identify that the first operation in the strided STDC module (a pointwise convolution) is the most expensive in terms of the number of multiply-accumulate (MAC) operations. By relocating this operation to the bottom path after the average pooling operation, we make the strided STDC block computationally more tractable overall while increasing the number of features in the top path and the number of features with a large receptive field size in the concatenated output. Our modified blocks lead to a reduction of over 50\% of the MAC operations in stage 1, and of over 10\% in stages 2 and 3.


\subsection{Reduced input/output dimensionality}
We design the network for low-resolution input and downscale all dataset's input frames, optical flow, and motion boundary ground truth accordingly. The scaling factor is picked such that the resulting data resolution closely matches the target application resolution (approx. qqVGA, 160x120 pixels). Horizontal and vertical scalings are identical, to fix the aspect ratio in an attempt to retain naturalism. This allows us to make the network shallower by dropping the first (expensive) convolution altogether and thus decrease latency while maintaining feature sizes in the deepest layers. The downscaled training data matches the low-resolution cameras found on nano quadcopters more closely, making our synthetic dataset more naturalistic for our intended application. As an added benefit, working with downscaled data significantly speeds up training. The primary downside of reduced input resolution is the loss of information, in particular we will miss out on small objects and small displacements that are not captured by the resolution. To be able to compare with existing optical flow works, we benchmark performance at native dataset resolution, since downscaling of flow magnitudes results in lower endpoint error (EPE) without a qualitative improvement. 


Lastly, we design our network for grayscale input images, saving two third of the on-board memory dedicated to the input frames and decreasing the computational cost of the first layer (at a loss of color information). 


\begin{figure*}[t]
\centering
\begin{subfigure}[t]{0.195\textwidth}
    \includegraphics[width=\textwidth] 
    {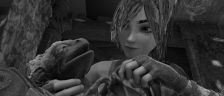}
    \includegraphics[width=\textwidth] 
    {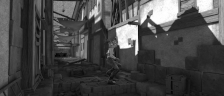}
    \caption{Input frame I}
\end{subfigure}\hfill
\begin{subfigure}[t]{0.195\textwidth}
    \includegraphics[width=\textwidth]  
    {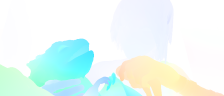}
    \includegraphics[width=\textwidth]
    {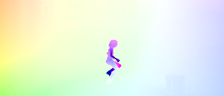}
    \caption{Ground truth}
\end{subfigure}\hfill
\begin{subfigure}[t]{0.195\textwidth}
    \includegraphics[width=\textwidth]  
    {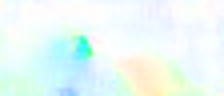}
    \includegraphics[width=\textwidth]
    {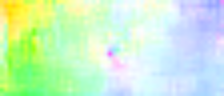}
    \caption{NanoFlowNet}
\end{subfigure}\hfill
\begin{subfigure}[t]{0.195\textwidth}
    \includegraphics[width=\textwidth]  
    {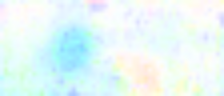}
    \includegraphics[width=\textwidth]
    {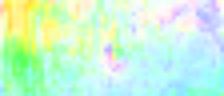}
    \caption{NanoFlowNet-s}
\end{subfigure}\hfill
\begin{subfigure}[t]{0.195\textwidth}
    \includegraphics[width=\textwidth]  
    {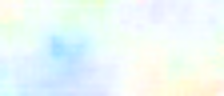}
    \includegraphics[width=\textwidth]
    {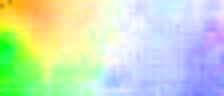}
    \caption{FlowNet2-xs}
\end{subfigure}
\caption{Qualitative comparison of optical flow estimates by NanoFlowNet(-s) and FlowNet2-xs on MPI Sintel (train) clean pass.} 
\label{fig:results}
\end{figure*}

\begin{table*}[t]
\centering
\begin{tabular}{llllll}
\hline
\multirow{2}{*}{\textbf{Method}} & \multicolumn{2}{l}{\textbf{MPI Sintel (train) [EPE]}} & \multicolumn{2}{l}{\textbf{Frame rate [FPS]}}       & \multirow{2}{*}{\textbf{Parameters}} \\ \cline{2-5}
                                 & \textit{\textbf{Clean}}   & \textit{\textbf{Final}}   & \textit{\textbf{GPU¹}} & \textit{\textbf{GAP8²}} &                                      \\ \hline
FlowNet2-xs           & 9.054                     & 9.458                     & 150           & -                     & 1,978,250                            \\
NanoFlowNet (ours)               & \textbf{7.122}            & \textbf{7.979}            & 141                    & 5.57                    & 170,881                              \\
NanoFlowNet-s (ours)             & 9.559                     & 10.047                    & \textbf{151}                    & \textbf{9.34}           & 46,749                               \\ \hline
\end{tabular}
\caption{Quantitative results on MPI Sintel. ¹At a resolution of 96x224. ²At a resolution of 112x160, including vision thread.}
\label{tab:results}
\end{table*}

\section{Experiments}

\subsection{Implementation details}
All models are trained for 300 epochs on FlyingChairs2 \cite{Dosovitskiy2015, Ilg2018a}, a regenerated FlyingChairs dataset with motion boundary ground truth. We use the Adam optimizer \cite{Kingma2014}, with learning rate 1e-3 and a batch size of 8. After this, we fine-tune our architectures on FlyingThings3D \cite{Mayer2016a} for 200 epochs with a learning rate of 1e-4. 

Given the scaling and conversion to grayscale of the input data, our network is not directly comparable with results reported by other works. For comparison, we retrain one of the fastest networks in literature, Flownet2-s \cite{Ilg2017}, on the same data. Given the reduction in resolution, we drop the deepest two layers to maintain a reasonable feature size in the deepest layers, and name the model \textit{Flownet2-xs}. 

We run all experiments in a docker environment with TensorFlow 2.8.0, CUDA 11.2, CUDNN 8.1.0, TensorRT 7.2.2 on an NVIDIA GeForce GTX 1070 Max-Q with batch size 1 for benchmarking latency.

\begin{table}[t]
\centering
\begin{tabular}{lll}
\hline
\multirow{2}{*}{\textbf{Detail guidance method}} & \multicolumn{2}{l}{\textbf{MPI Sintel (train)}}   \\ \cline{2-3} 
                                                 & \textit{\textbf{Clean}} & \textit{\textbf{Final}} \\ \hline
None                                             & 7.636                   & 8.119                   \\
Edge detect                                      & 7.404                   & 8.141                   \\
Motion boundaries                                & \textbf{7.122}          & \textbf{7.979}          \\ \hline
\end{tabular}
\caption{Quantitative comparison of different methods of detail guidance.}
\label{tab:ablation_detail_guidance}
\end{table}

\subsection{Performance and latency on public benchmarks}
We evaluate the trained networks on the MPI Sintel train subset, on both the clean and final pass. Quantitative results can be found in Table \ref{tab:results}. Regarding accuracy, according to these results, our NanoFlowNet performs better than the squeezed FlowNet2-xs architecture, despite using less than 10\% of the parameters. With respect to runtime, FlowNet2-xs does not fit on the GAP8 microprocessor due to the network size (i.e., lack of memory). To put the achieved latency of NanoFlowNet in perspective, we execute FlowNet2-xs' first two convolutions and the final prediction layer on the GAP8. The three-layer architecture achieves 4.96 FPS, which is slower than running the entire NanoFlowNet (5.57 FPS). On laptop GPU hardware, NanoFlowNet achieves comparable FPS to FlowNet2-xs. NanoFlowNet-s has lower performance than both other models, but has a low parameter count with only 27\% of NanoFlowNet's and 2.4\% of FlowNet2-xs's parameters, and is the fastest out of all the networks tested.

\begin{figure*}[t]
\centering
\begin{subfigure}[t]{0.195\textwidth}
    \includegraphics[width=\textwidth] 
    {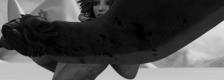}
    \includegraphics[width=\textwidth] 
    {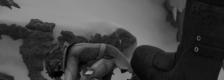}
    \includegraphics[width=\textwidth] 
    {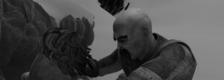}
    \caption{Input frame I}
\end{subfigure}\hfill
\begin{subfigure}[t]{0.195\textwidth}
    \includegraphics[width=\textwidth]  
    {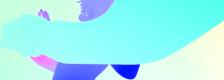}
    \includegraphics[width=\textwidth]
    {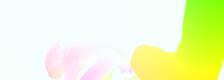}
    \includegraphics[width=\textwidth]
    {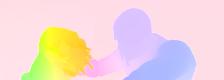}
    \caption{Ground truth}
\end{subfigure}\hfill
\begin{subfigure}[t]{0.195\textwidth}
    \includegraphics[width=\textwidth]  
    {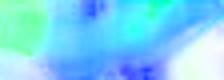}
    \includegraphics[width=\textwidth]
    {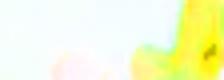}
    \includegraphics[width=\textwidth]
    {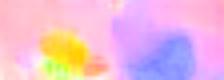}
    \caption{Motion-boundary guidance}
\end{subfigure}\hfill
\begin{subfigure}[t]{0.195\textwidth}
    \includegraphics[width=\textwidth]  
    {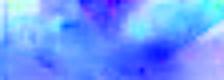}
    \includegraphics[width=\textwidth]
    {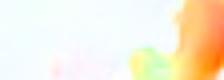}
    \includegraphics[width=\textwidth]
    {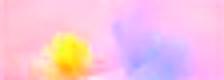}
    \caption{Edge-detect guidance}
\end{subfigure}\hfill
\begin{subfigure}[t]{0.195\textwidth}
    \includegraphics[width=\textwidth]  
    {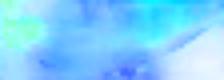}
    \includegraphics[width=\textwidth]
    {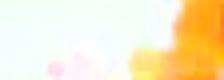}
    \includegraphics[width=\textwidth]
    {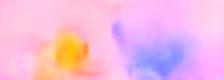}
    \caption{No detail guidance}
\end{subfigure}
\caption{Qualitative comparison of different detail guidance methods on MPI Sintel (train) clean pass.}
\label{fig:qualitative_guidance}
\end{figure*}

Qualitative results, presented in Fig. \ref{fig:results}, confirm that NanoFlowNet makes the most accurate optical flow estimates out of all the networks tested. Interestingly, both NanoFlowNet and NanoFlowNet-s appear to detect displacements of smaller objects, which FlowNet2-xs misses. However, NanoFlowNet-s' flow estimates are highly noisy.


\subsection{Ablation study}
\subsubsection{Motion boundaries detail guidance}
We verify the effectiveness of motion boundary detail guidance by retraining two additional networks, one with detail guidance based on the optical flow ground truth convolved with a Laplacian kernel (further referred to as ``edge-detect guidance''), and another one with no detail guidance. Quantitative and qualitative results can be found in Table \ref{tab:ablation_detail_guidance} and Fig. \ref{fig:qualitative_guidance}, respectively. As shown, motion boundary detail guidance improves results and outperforms edge detect detail guidance. Since all these guidance methods only affect (i.e., guide) the training behavior, all methods have identical latency. Qualitative results show that motion-boundary-guided optical flow best defines moving objects, and shows the least ``leakage'' of foreground objects into the background.


\subsubsection{Strided STDC module redesign}
Table \ref{tab:strided_stdc} shows the effects of the strided STDC module redesign. The network with the redesigned module is both faster (both on laptop GPU and the GAP8 microprocessor) and more accurate.

\begin{table}[t]
\centering
\begin{tabular}{lllll}
\hline
\multirow{2}{*}{\textbf{Strided STDC block}} & \multicolumn{2}{l}{\textbf{MPI Sintel (train) [EPE]}} & \multicolumn{2}{l}{\textbf{Frame rate [FPS]}} \\ \cline{2-5} 
                                             & \textit{\textbf{Clean}}   & \textit{\textbf{Final}}   & \textbf{\textit{GPU¹}}             & \textbf{\textit{GAP8²}}            \\ \hline
Unmodified                                   & 7.483                     & 8.114                     &  136               & 4.84                 \\
Modified                                     & \textbf{7.122}            & \textbf{7.979}            & \textbf{141}    & \textbf{5.57}  
\end{tabular}
\caption{Quantitative comparison of the original and the modified strided STDC block. ¹At a resolution of 96x224. ²At a resolution of 112x160, including vision thread.}
\label{tab:strided_stdc}
\end{table}

\subsubsection{Reduced input dimensionality}

A comparison between training and inferring on grayscale images compared to color images can be found in Table \ref{tab:color}. Our grayscale model outperforms the color variant. We hypothesize that this is due to the limited capacity of the network. The latency of the grayscale model on the GAP8 is lower due to reduced data transfer and a cheaper first convolution.

\begin{table}[t]
\centering
\begin{tabular}{lllll}
\hline
\multirow{2}{*}{\textbf{Mode}} & \multicolumn{2}{l}{\textbf{MPI Sintel (train) [EPE]}} & \multicolumn{2}{l}{Frame rate [FPS]}                \\ \cline{2-5} 
                                             & \textit{\textbf{Clean}}   & \textit{\textbf{Final}}   & \textbf{\textit{GPU¹}} & \textbf{\textit{GAP8²}} \\ \hline
Color                                        & 7.726                     & 8.344                     & \textbf{141}           & 5.18                        \\
Grayscale                                    & \textbf{7.122}            & \textbf{7.979}            & \textbf{141}           & \textbf{5.57}          
\end{tabular}
\caption{Quantitative comparison of grayscale vs. color input frame-based architectures. ¹At a resolution of 96x224. ²At a resolution of 112x160, including vision thread.}
\label{tab:color}
\end{table}


\subsection{Obstacle avoidance implementation}

\begin{figure}[t]
    \centering
    \includegraphics[width=0.35\textwidth]{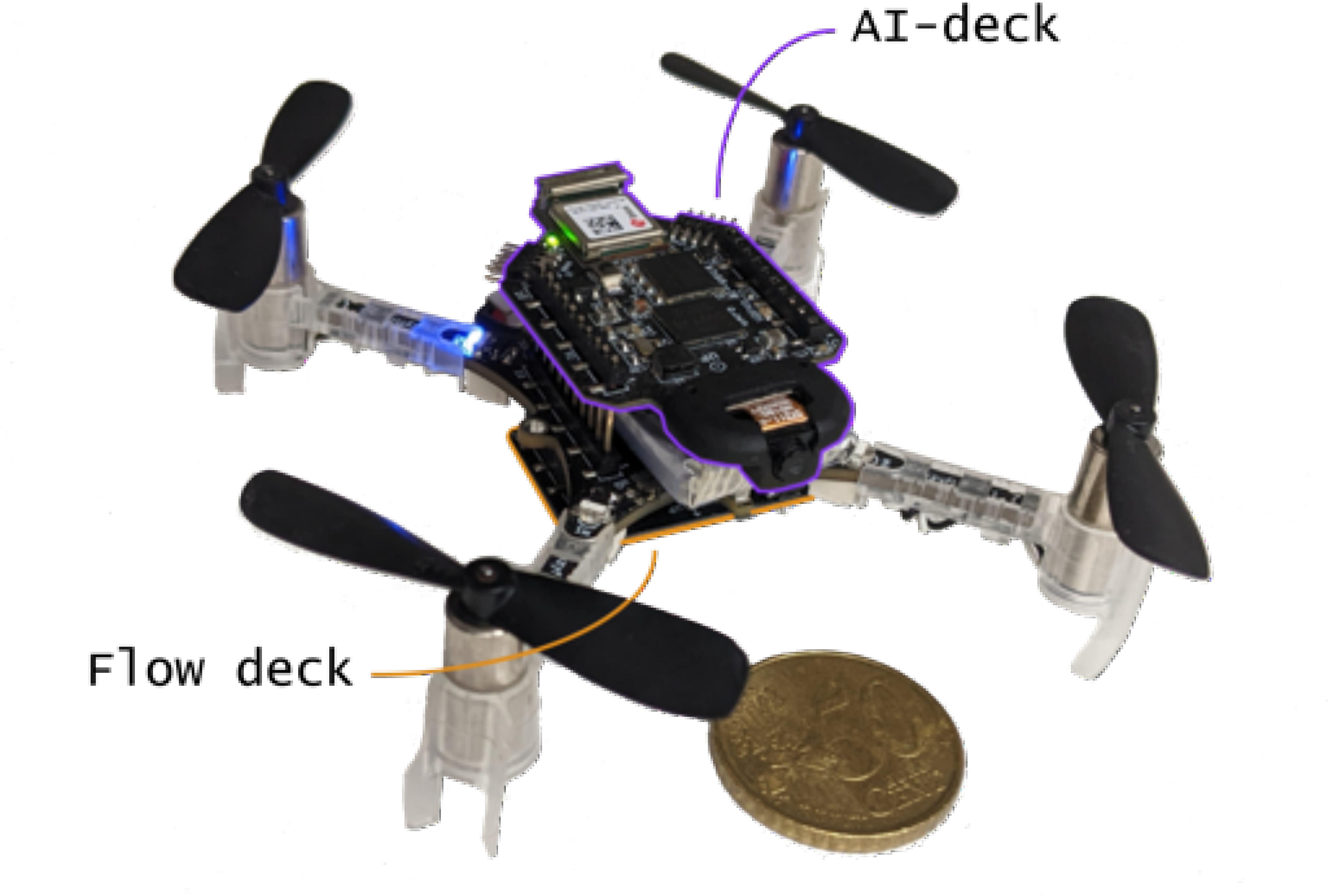}
    \caption{Crazyflie 2.x equipped with (i) the AI-deck used for image acquisition using a front-facing camera and to run optical flow inference, and (ii) the downward-facing flow-deck used only for positioning.}
    \label{fig:obstacleAvoidance}
\end{figure}

We deploy the proposed NanoFlowNet architecture on a Crazyflie 2.x equipped with the AI-deck and the flow-deck for the task of vision-based obstacle avoidance. We use the AI-deck to capture images with the front-facing camera and to run optical flow inference and processing. The downward-facing optical flow deck is used for positioning only. The total flight platform weighs in at 34 g. See Fig. \ref{fig:obstacleAvoidance} for a picture of the platform.

\subsubsection{Control strategy}

We implement the horizontal balance strategy from \cite{Souhila2007, Cho2019}, with which 
the yaw rate $\dot{\psi}$ is set based on the error $e_{rl}$ between the sum of flow magnitudes in the left and right half of the flow estimate 
(see Eq. \ref{eq:yaw_rate}). We set gains $k_p=0.0126$ and $k_d=0.0018$ experimentally. 
The forward velocity of the quadcopter is set at 0.2 m/s.
\begin{equation}\label{eq:yaw_rate}
    \dot{\psi} = k_p e_{rl} + k_d \dot{e_{rl}}
\end{equation}

\begin{figure}
\centering
    \includegraphics[width=0.45\textwidth] 
    {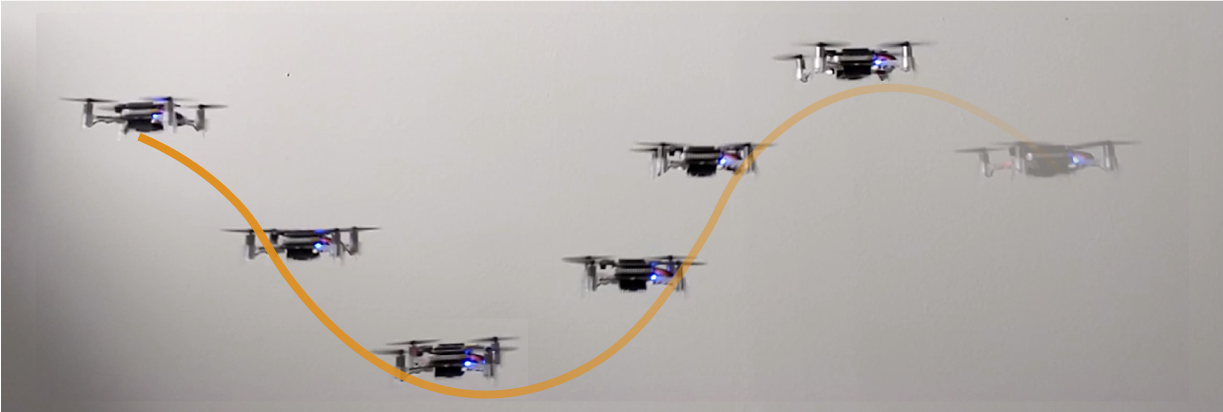}
\caption{Inspired by GapFlyt \cite{Sanket2018}, we deliberately let the quadcopter oscillate vertically to generate additional optical flow.}
\label{fig:pipeline}
\end{figure}

We augment the balance strategy by implementing active oscillations (a cyclic up-down movement, see Fig. \ref{fig:pipeline}) which results in additional optical flow being generated across the field of view (FOV). This is particularly helpful for avoiding objects in the direction of horizontal travel. Up-down rather than left-right surveying favors detecting obstacles wider than taller in nature, but is much simpler to combine with the left-right balance strategy. Additionally, left-right surveying requires rolling, which introduces rotational flow that does not contain depth information.

We implement both the CNN and calculation of $e_{rl}$ on the GAP8 microprocessor of the AI-deck. Calculating the flow error on the AI-deck significantly reduces the amount of data that needs to be transmitted over UART to the autopilot. The calculation of the yaw rate is done on the Crazyflie 2.x, and fed into the controller.

\subsubsection{AI-deck implementation}
The CNN processing power on the AI-deck comes from the GreenWaves Technologies GAP8. The chip is organized around the central single-core fabric controller (FC) and the eight-core cluster (CL) for parallelized workloads. For our application, we run FC@250MHz, CL@230MHz, and VDD@1.2V.

Our AI-deck is equipped with the HM01B0 monochrome camera, which supports a resolution of up to 324x324, a QVGA (244x324) window mode, a 2x2 monochrome binning mode, and cropping. For our application we enable both the window mode and binning mode (122x162) and take a central crop of 112x160, to ensure a matched spatial resolution of upsampled and skipped features in the network architecture. At our input resolution, using grayscale versus color reduces the L2 memory usage on the AI-deck by 14\%. This additional L2 memory is made available to the AutoTiler, which improves inference time by reducing the number of data transfers.

In this work, we utilize the GreenWaves Technologies GAPflow toolset for porting our CNN to the GAP8. NNTool takes a TensorFlow Lite or ONNX CNN description and maps all operations and parameters to a representation compatible with AutoTiler, the GAPflow tiling solver.

We use NNTool to implement 8-bit post-training quantization to our CNN. We quantize on images from the MPI Sintel dataset \cite{Butler2012} and achieve an average signal to quantization noise ratio (SQNR) of 10.

\begin{figure*}
\centering
\begin{subfigure}[t]{0.275\textwidth}
    \includegraphics[height=0.185\textheight, width=0.191\textheight] 
    {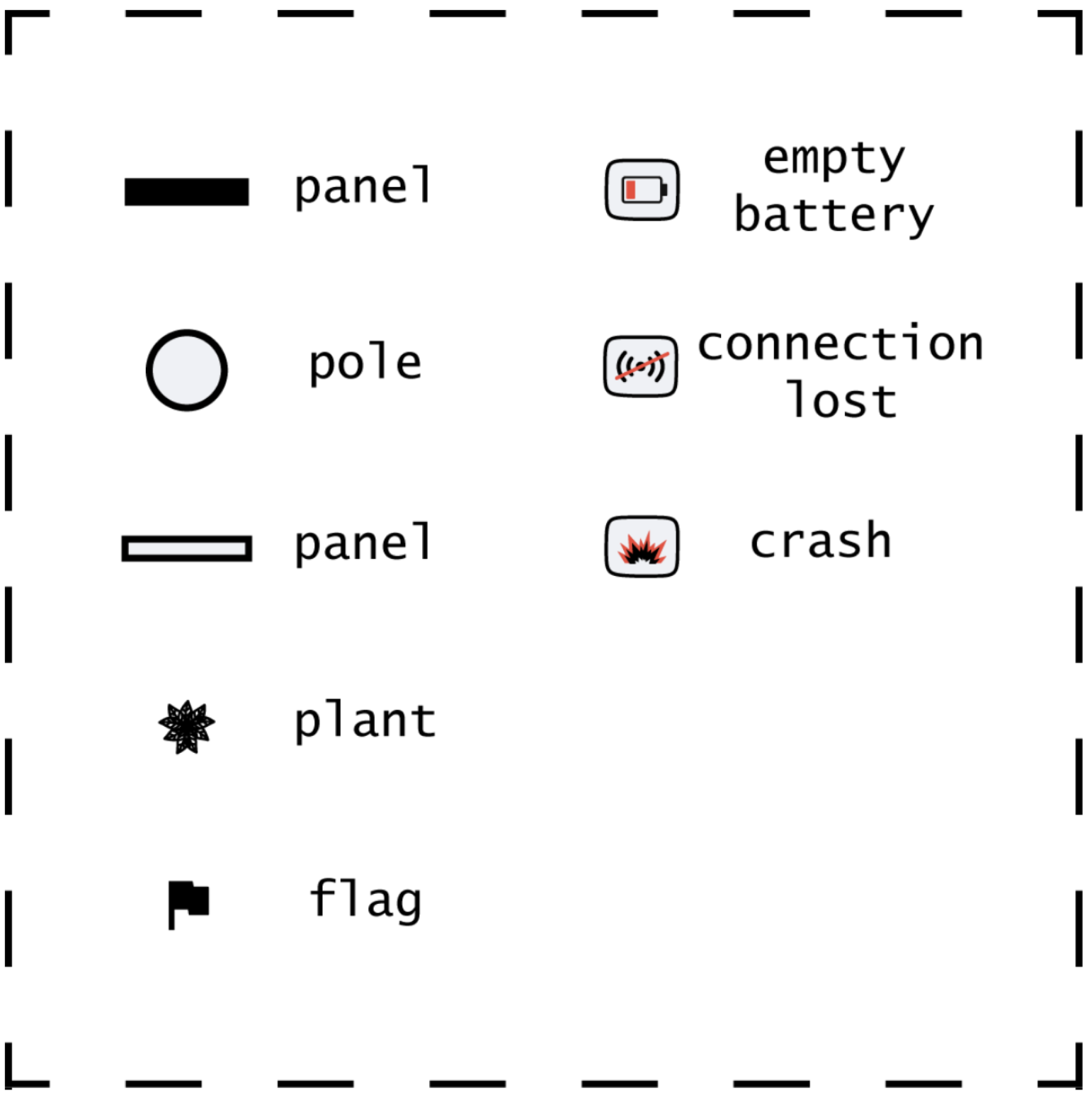}
\end{subfigure}\hspace{5pt}
\begin{subfigure}[t]{0.275\textwidth}
    \includegraphics[height=0.185\textheight]  
    {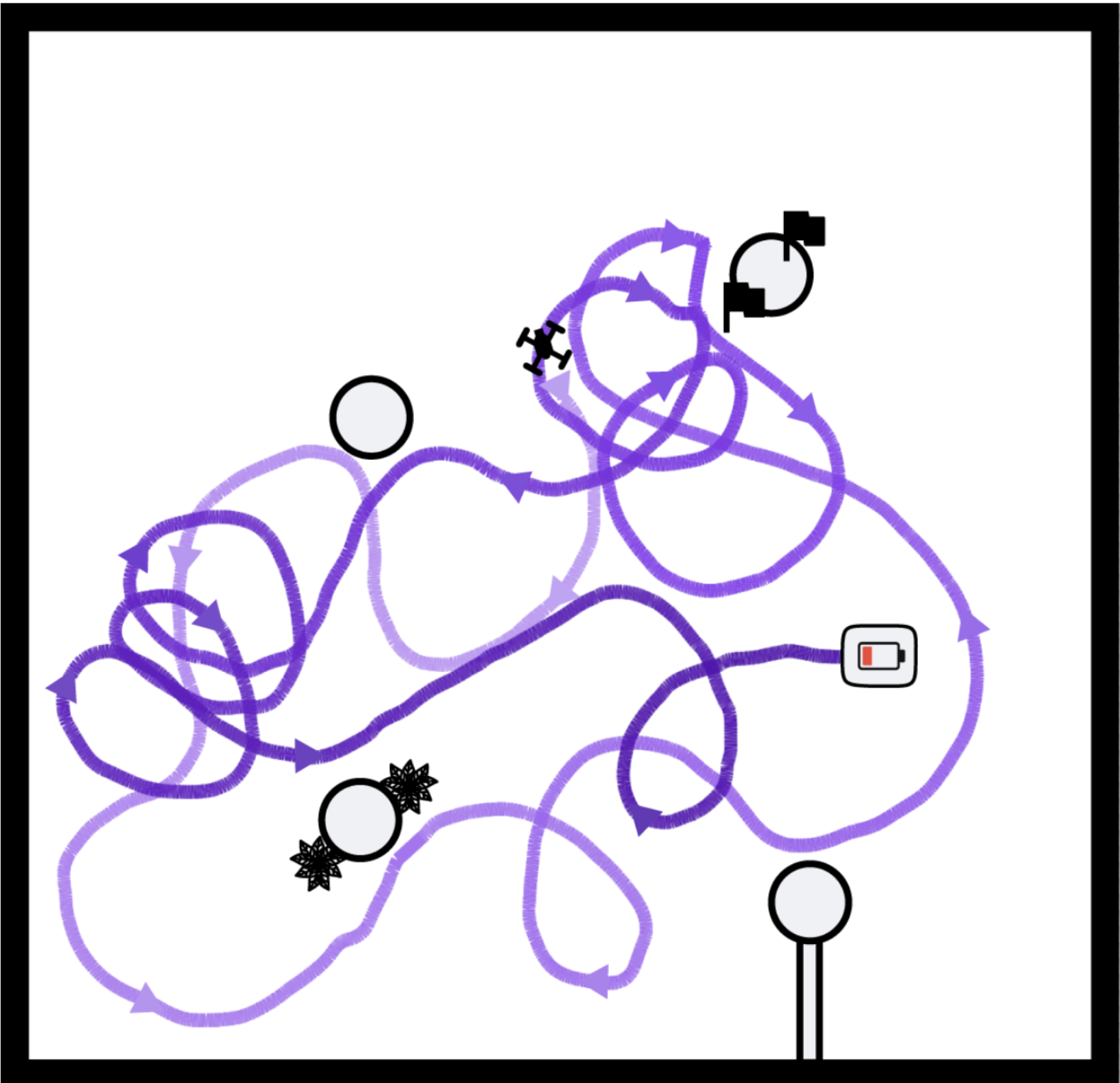}
\end{subfigure}\hspace{5pt}
\begin{subfigure}[t]{0.275\textwidth}
    \includegraphics[height=0.185\textheight, width=0.191\textheight]  
    {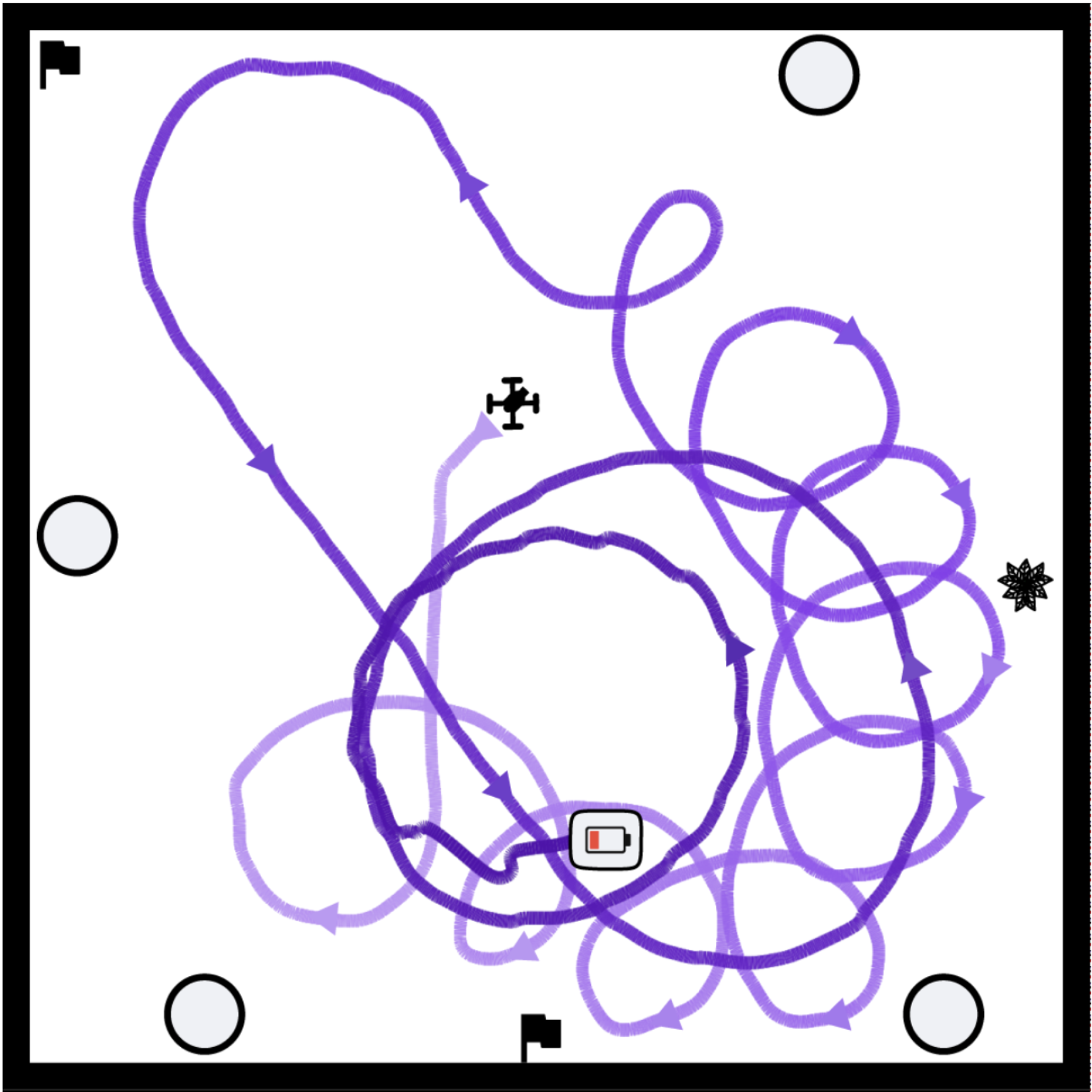}
\end{subfigure}\\\vspace{5pt}
\begin{subfigure}[t]{0.275\textwidth}
    \includegraphics[height=0.185\textheight]  
    {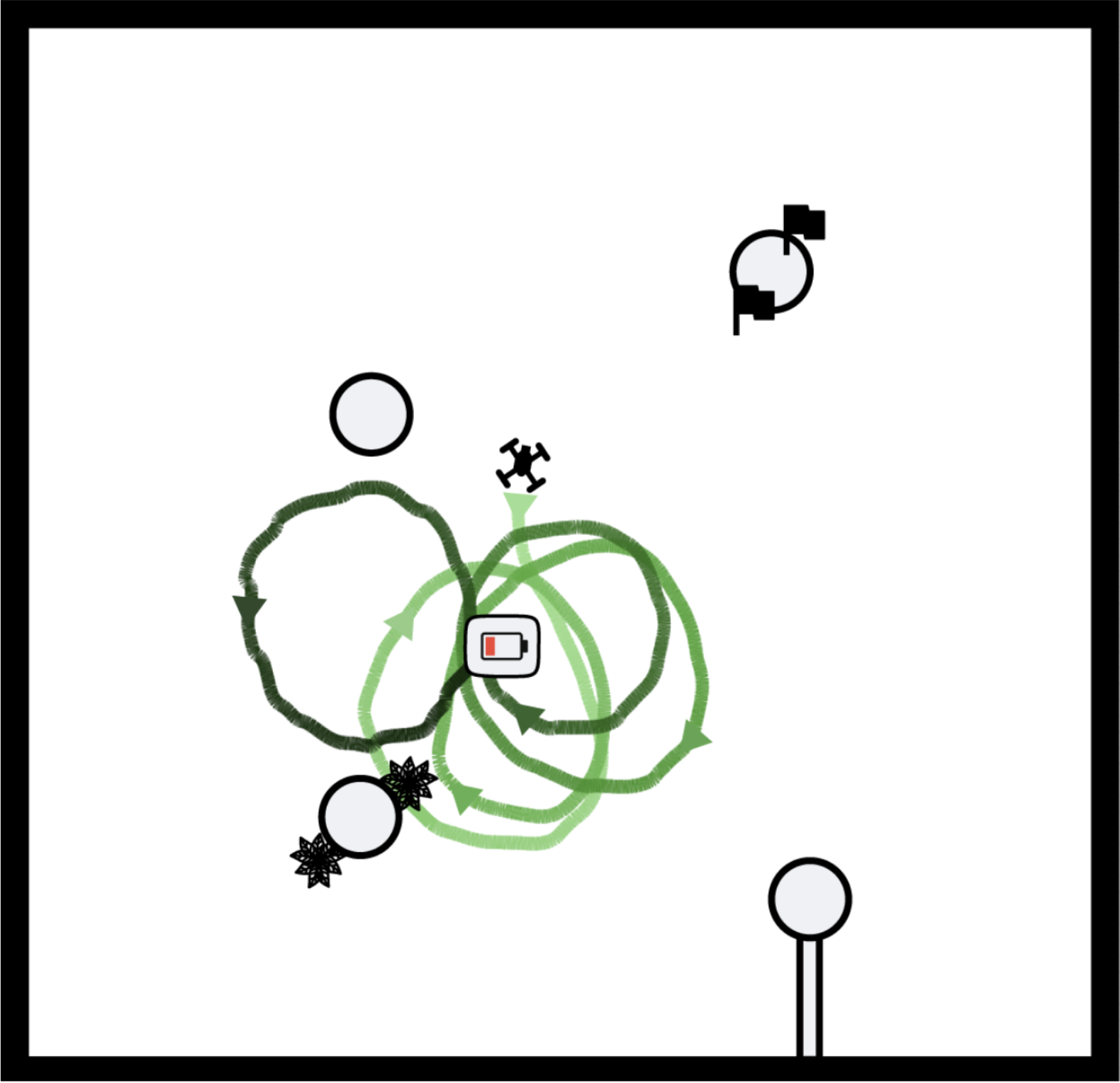}
\end{subfigure}\hspace{5pt}
\begin{subfigure}[t]{0.275\textwidth}
    \includegraphics[height=0.185\textheight] 
    {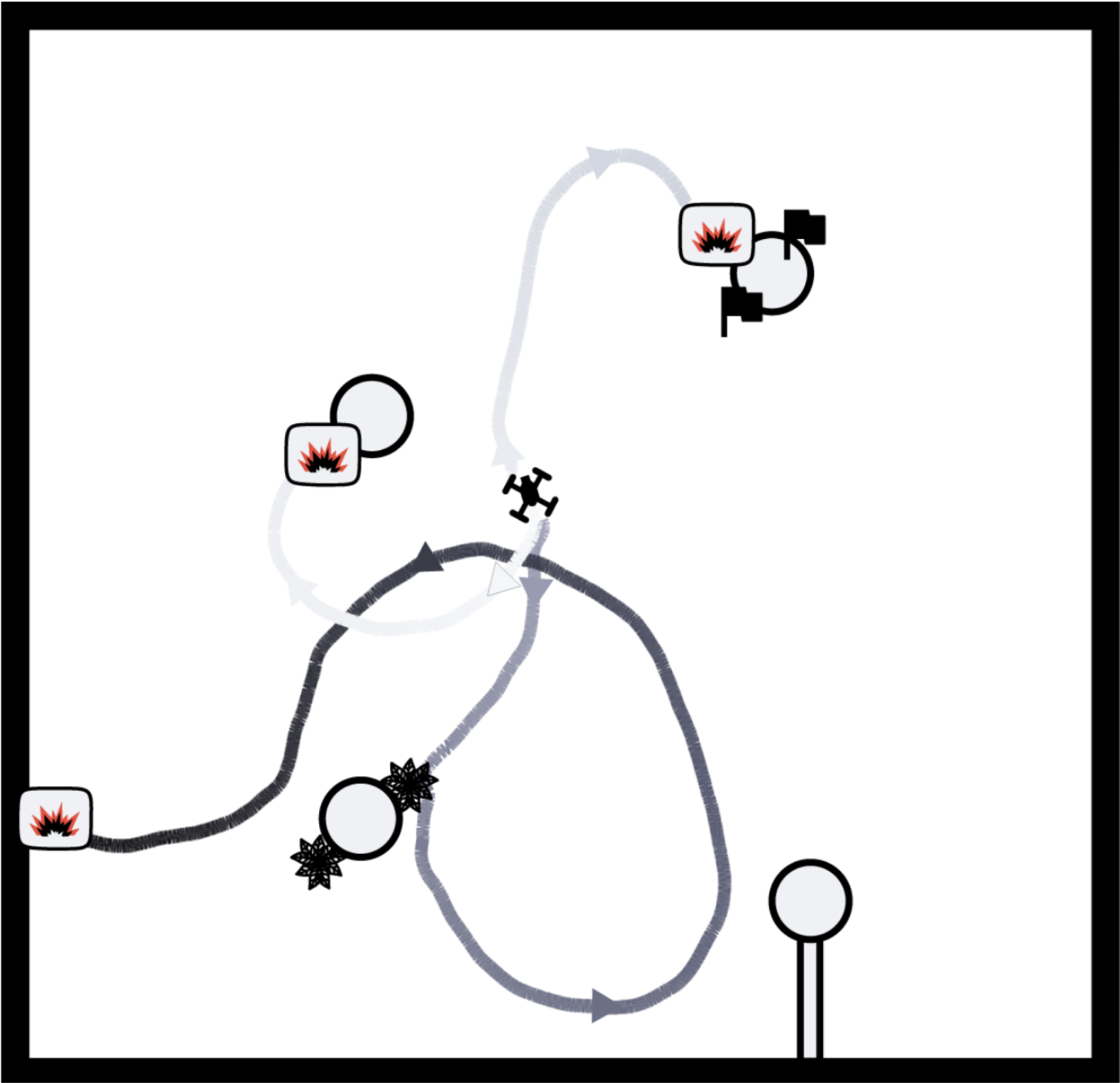}
\end{subfigure}\hspace{5pt}
\begin{subfigure}[t]{0.275\textwidth}
    \includegraphics[height=0.185\textheight]  
    {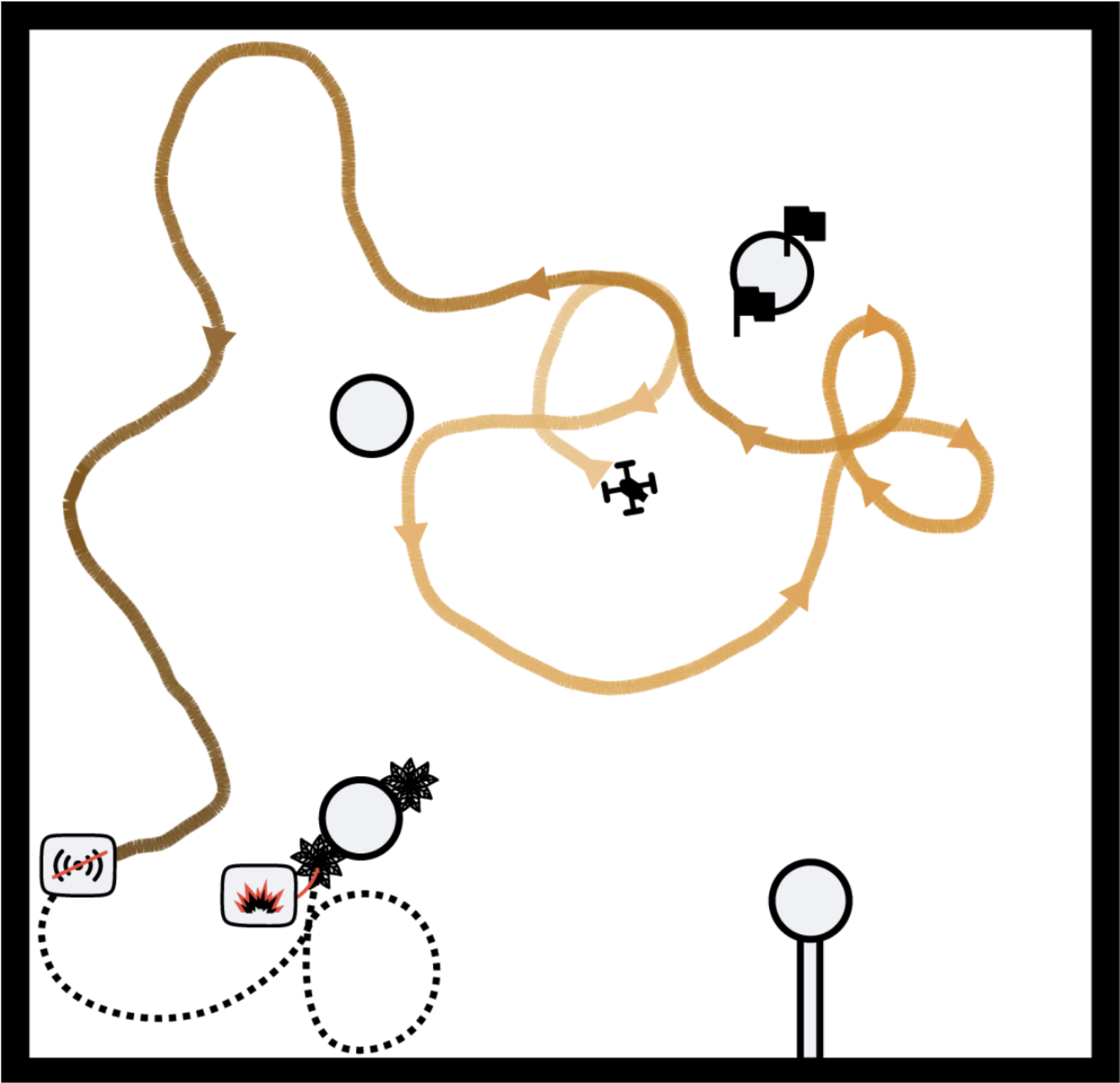}
\end{subfigure}
\caption{Results of multiple obstacle avoidance runs in cluttered and open environments. Position recorded with an OptiTrack Motion Capture System.}
\label{fig:avoidanceresults}
\end{figure*}

\subsubsection{Experimental setup}
\begin{figure}
\centering
\begin{subfigure}[t]{0.4\textwidth}
    \includegraphics[height=0.225\textheight]  
    {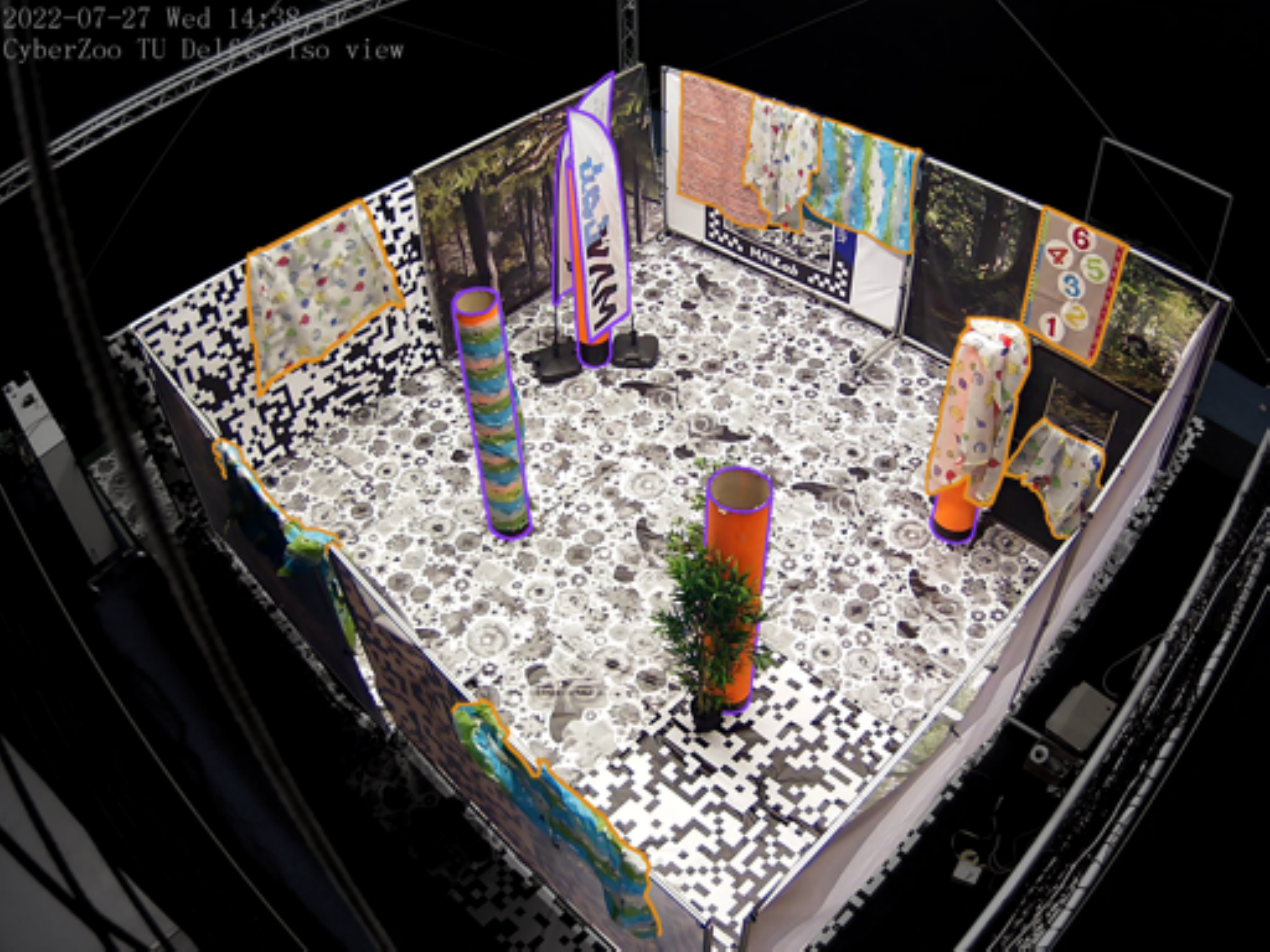}
\end{subfigure}\hfill\vfill
\caption{Overview of the cluttered, obstacle avoidance environment. Obstacles are outlined in purple, while texture-enhancing mats and curtains in orange.}
\label{fig:cyberzoo}
\end{figure}

We compose two indoor environments for obstacle avoidance. First, an open environment, with obstacles exclusively placed at the outline of the environment. Second, a cluttered environment, with obstacles placed throughout (see Fig. \ref{fig:cyberzoo}). Obstacles include textured and untextured poles, synthetic plants, flags, or panels. Both environments are enclosed with textured panels to trap the quadcopter inside. Panel textures consist of forest texture, data matrix texture, and a drone racing gate texture. In both environments, we augment the enclosure's texture with highly textured mats and curtains.

The simple proof-of-concept control algorithm has no dedicated method of dealing with head-on collisions. By placing obstacles around the perimeter of the open environment we minimize the risk of a head-on collision with the panels as they introduce an imbalance of optical flow, even on a fully perpendicular collision path with a panel.

For each experiment, we start the quadcopter at approximately the same location, with varying heading. We let the quadcopter run until a collision or empty battery. We record flight positioning data with an OptiTrack Motion Capture System for post-flight analysis only and record experiments with an ISO view and top view camera.

\subsubsection{Results}
Flight paths extracted from the motion capture system are plotted on maps of the environment and can be found in Fig. \ref{fig:avoidanceresults}. The control algorithm is most robust in the open environment, with the quadcopter managing to drain a full battery without crashing. In the cluttered environment, performance is much more variable. Especially in occasions where obstacles are in close proximity to one another, the quadcopter tends to successfully avoid an obstacle, only to turn straight into another. Adding a head-on collision detection based on the detection of the focus-of-expansion (FOE) and divergence estimation (e.g., \cite{DeCroon2021}) could help avoid obstacles in these cases. 

In several of the successful avoidances, the quadcopter initially responds weakly to the obstacle, only to turn away more harshly when the course has already been corrected sufficiently. This behavior is expected because of two reasons. First, the optical flow due to forward movement is zero at the FOE and maximum at the edge of the peripheral vision. 
Second, due to the fact that the obstacles take up more of the FOV when they are in closer proximity to the quadcopter, they generate more optical flow. This behavior could be corrected by weighing the optical flow more heavily towards the center of the image.

Another notable feature of the flight paths is that the nano quadcopter frequently appears to enter a spiraling path. The control algorithm is overreacting to stimuli from across the environment. Despite this, the behavior is consistent, the resulting paths are still exploring the environments, and the nano quadcopter is able to break out of the spiraling motion by approaching a panel (see Fig. \ref{fig:avoidanceresults}, top right) or approaching an obstacle (see Fig. \ref{fig:avoidanceresults}, top center). 

\section{Conclusions \& Discussion}
\label{discussion}

In this work, we introduced a lightweight CNN architecture for dense optical flow estimation on edge hardware, called NanoFlowNet. We achieved real-time latency on the AI-deck. Furthermore, we showed that training our network guided on motion boundaries improves performance at zero cost to latency. Finally, we implemented NanoFlowNet in a real-world obstacle avoidance application on board a Bitcraze Crazyflie nano quadcopter. For future work, we expect examples that take more advantage of the dense information in the generated optical flow field.




\end{document}